\documentclass[sigconf]{acmart}

\usepackage{booktabs} 

\usepackage{amssymb}
\usepackage{amsthm, amsmath}
\usepackage{algorithmic}
\usepackage[ruled]{algorithm2e}
\usepackage{color}
\usepackage{url}
\usepackage{xfrac}

\usepackage{subfigure}
\usepackage{tablefootnote}

\usepackage{tabularx}
\usepackage{multirow}
\usepackage{balance}

\begin{document}
\title{Matching Natural Language Sentences \\with Hierarchical Sentence Factorization}

\copyrightyear{2018}
\acmYear{2018} 
\setcopyright{iw3c2w3}
\acmConference[WWW 2018]{The 2018 Web Conference}{April 23--27, 2018}{Lyon, France}
\acmBooktitle{WWW 2018: The 2018 Web Conference, April 23--27, 2018, Lyon, France}
\acmPrice{}
\acmDOI{10.1145/3178876.3186022}
\acmISBN{978-1-4503-5639-8}

\author{Bang Liu$^1$, Ting Zhang$^1$, Fred X. Han$^1$, Di Niu$^1$, Kunfeng Lai$^2$, Yu Xu$^2$}
       \affiliation{$^1$University of Alberta, Edmonton, AB, Canada}
 \affiliation{$^2$Mobile Internet Group, Tencent, Shenzhen, China}



\begin{abstract}
Semantic matching of natural language sentences or identifying the relationship between two sentences is a core research problem underlying many natural language tasks.
Depending on whether training data is available, prior research has proposed both unsupervised distance-based schemes and supervised deep learning schemes for sentence matching.
However, previous approaches either omit or fail to fully utilize the ordered, hierarchical, and flexible structures of language objects, as well as the interactions between them.
In this paper, we propose \textit{Hierarchical Sentence Factorization}---a technique 
to factorize a sentence into a hierarchical representation, with the components at each different scale reordered into a ``predicate-argument'' form. The proposed sentence factorization technique leads to the invention of: 1) a new unsupervised distance metric which calculates the semantic distance between a pair of text snippets by solving a penalized optimal transport problem while preserving the logical relationship of words in the reordered sentences, and 2) new multi-scale deep learning models for supervised semantic training, based on factorized sentence hierarchies.
We apply our techniques to text-pair similarity estimation and text-pair relationship classification tasks, based on multiple datasets such as STSbenchmark, the Microsoft Research paraphrase identification (MSRP) dataset, the SICK dataset, etc. Extensive experiments show that the proposed hierarchical sentence factorization can be used to significantly improve the performance of existing unsupervised distance-based metrics as well as multiple supervised deep learning models based on the convolutional neural network (CNN) and long short-term memory (LSTM).
\end{abstract}

%
%



\settopmatter{printacmref=false}

\maketitle

\section{Introduction}
\label{sec:intro}

Semantic matching, which aims to model the underlying semantic similarity or dissimilarity among different textual elements such as sentences and documents, has been playing a central role in many Natural Language Processing (NLP) applications, including information extraction \cite{grishman1997information}, top-$k$ re-ranking in machine translation \cite{brown1993mathematics}, question-answering \cite{yu2014deep}, automatic text summarization \cite{ponzanelli2015summarizing}.
However, semantic matching based on either supervised or unsupervised learning remains a hard problem. Natural language demonstrates complicated hierarchical structures, where different words can be organized in different orders to express the same idea. As a result, appropriate semantic representation of text plays a critical role in matching natural language sentences.

Traditional approaches represent text objects as bag-of-words (BoW), term frequency inverse document frequency (TF-IDF) \cite{wu2008interpreting} vectors, or their enhanced variants \cite{paltoglou2010study,robertson1994some}. However, such representations can not accurately capture the similarity between individual words, and do not take the semantic structure of language into consideration. 
Alternatively, word embedding models, such as \textit{word2vec} \cite{mikolov2013efficient} and \textit{Glove} \cite{pennington2014glove}, learn a distributional semantic representation of each word and have been widely used. 

Based on the word-vector representation, a number of unsupervised and supervised matching schemes have been recently proposed.
As an unsupervised learning approach, the Word Mover's Distance (WMD) metric \cite{kusner2015word} measures the dissimilarity between two sentences (or documents) as the minimum distance to transport the embedded words of one sentence to those of another sentence. However, the sequential and structural nature of sentences is omitted in WMD. For example, two sentences containing  exactly the same words in different orders can express totally different meanings.
On the other hand, many supervised learning schemes based on deep neural networks have also been proposed for sentence matching \cite{mueller2016siamese,severyn2015learning,wang2017bilateral,pang2016text}. A common characteristic of many of these neural network models is that they adopt a Siamese architecture, taking the word embedding sequences of a pair of sentences (or documents) as the input, transforming them into intermediate contextual representations via either convolutional or recurrent neural networks, and performing scoring over the contextual representations to yield final matching results. However, these methods rely purely on neural networks to learn the complicated relationships among sentences, and many obvious compositional and hierarchical features are often overlooked or not explicitly utilized.

In this paper, however, we argue that a successful semantic matching algorithm needs to best characterize the sequential, hierarchical and flexible structure of natural language sentences, as well as the rich interaction patterns among semantic units.
We present a technique named \textit{Hierarchical Sentence Factorization} (or \emph{Sentence Factorization} in short), which is able to represent a sentence in a hierarchical semantic tree, with each node (semantic unit) at different depths of the tree reorganized into a normalized ``predicate-argument'' form.
Such normalized sentence representation enables us to propose new methods to both improve unsupervised semantic matching by taking the structural and sequential differences between two text entities into account, and enhance a range of supervised semantic matching schemes, by overcoming the limitation of the representation capability of convolutional or recurrent neural networks, especially when labelled training data is limited.
Specifically, we make the following contributions:


\textit{First}, the proposed \textit{Sentence Factorization} scheme factorizes a sentence recursively into a hierarchical tree of semantic units, where each unit is a subset of words from the original sentence.
Words are then reordered into a ``predicate-argument'' structure.
Such form of sentence representation offers two benefits: i) the flexible syntax structures of the same sentence, for example, active and passive sentences, can be normalized into a unified representation; ii) the semantic units in a pair of sentences can be aligned according to their depth and order in the factorization tree.

\textit{Second}, for unsupervised text matching, we combine the factorized and reordered representation of sentences and the Order-preserving Wasserstein Distance \cite{su2017order} (which was originally proposed to match hand-written characters in computer vision) to propose a new semantic distance metric between text objects, which we call \textit{Ordered Word Mover's Distance}. Compared with the recently proposed Word Mover's Distance  \cite{kusner2015word}, our new metric achieves significant improvement by taking the sequential structures of sentences into account. For example, without considering the order of words, the Word Mover's Distance between the sentences
 ``Tom is chasing Jerry'' and ``Jerry is chasing Tom'' is zero. In contrast, our new metric is able to penalize such order mismatch between words, and identify the difference between the two sentences.

\textit{Third}, for supervised semantic matching, we extend the existing Siamese network architectures (both for CNN and LSTM) to multi-scaled models, where each scale adopts an individual Siamese network, taking as input the vector representations of the two sentences at the corresponding depth in the factorization trees, ranging from the coarse-grained scale to fine-grained scales. 
When increasing the number of layers in the corresponding neural network can hardly improve performance, hierarchical sentence factorization provides a novel means to extend the original deep networks to a ``richer'' model that matches a pair of sentences through a multi-scaled semantic unit matching process. 
Our proposed multi-scaled deep neural networks can effectively improve existing deep models by measuring the similarity between a pair of sentences at different semantic granularities. For instance, Siamese networks based on CNN and BiLSTM \cite{mueller2016siamese,shao2017hcti} that originally only take the word sequences as the inputs.

We extensively evaluate the performance of our proposed approaches on the task of semantic textual similarity estimation and paraphrase identification, based on multiple datasets, including the STSbenchmark dataset, the Microsoft Research Paraphrase identification (MSRP) dataset, the SICK dataset and the MSRvid dataset. Experimental results have shown that our proposed algorithms and models can achieve significant improvement compared with multiple existing unsupervised text distance metrics, such as the Word Mover's Distance \cite{kusner2015word}, as well as supervised deep neural network models, including Siamese Neural Network models based on CNN and BiLSTM \cite{mueller2016siamese,shao2017hcti}.

The remainder of this paper is organized as follows.
Sec.~\ref{sec:sentence} presents our hierarchical sentence factorization algorithm.
Sec.~\ref{sec:owmd} presents our Ordered Word Mover's Distance metric based on sentence structural reordering.
In Sec.~\ref{sec:multi-layer}, we propose our multi-scaled deep neural network architectures based on hierarchical sentence representation.
In Sec.~\ref{sec:eval}, we conduct extensive evaluations of the proposed methods based on multiple datasets on multiple tasks.
Sec.~\ref{sec:related} reviews the related literature.
The paper is concluded in Sec.~\ref{sec:conclude}.

\section{Hierarchical Sentence Factorization and Reordering}
\label{sec:sentence}

\begin{figure*}[tb]
\centering
\includegraphics[width=\textwidth]{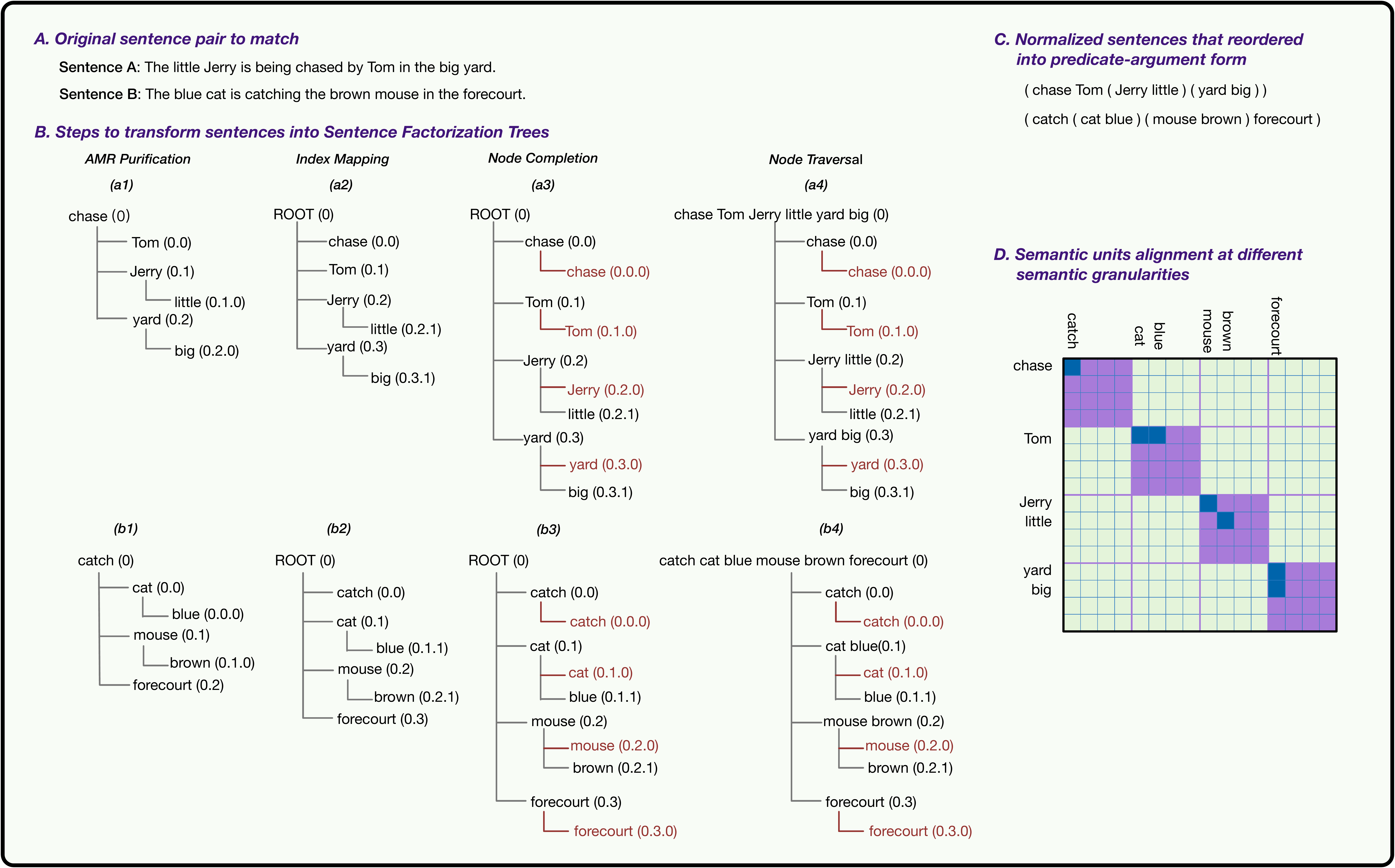}
\vspace{-3mm}
\caption{An example of the sentence factorization process. Here we show: A. The original sentence pair; B. The procedures of creating sentence factorization trees; C. The predicate-argument form of original sentence pair; D. The alignment of semantic units with the reordered form.}
\label{fig:casestudy}
\vspace{-1mm}
\end{figure*}

In this section, we present our \textit{Hierarchical Sentence Factorization} techniques to transform a sentence into a hierarchical tree structure, which also naturally produces a reordering of the sentence at the root node. 
This multi-scaled representation form proves to be effective at improving both unsupervised and supervised semantic matching, which will be discussed in Sec.~\ref{sec:owmd} and Sec.~\ref{sec:multi-layer}, respectively. 

We first describe our desired factorization tree structure before presenting the steps to obtain it. Given a natural language sentence $S$, our objective is to transform it into a semantic factorization tree denoted by $T^f_S$. Each node in $T^f_S$ is called a \textit{semantic unit}, which contains one or a few tokens (tokenized words) from the original sentence $S$, as illustrated in Fig.~\ref{fig:casestudy} $(a4)$, $(b4)$.
The tokens in every semantic unit in $T^f_S$ is re-organized into a ``predicate-argument'' form. For example, a semantic unit for ``Tom catches Jerry'' in the ``predicate-argument'' form will be ``catch Tom Jerry''.

Our proposed factorization tree recursively factorizes a sentence into a hierarchy of semantic units at different granularities to represent the semantic structure of that sentence.
The root node in a factorization tree contains the entire sentence reordered in the predicate-argument form, thus providing a ``normalized'' representation for sentences expressed in different ways (e.g., passive vs. active tenses). Moreover, each semantic unit at depth $d$ will be further split into several child nodes at depth $d + 1$, which are smaller semantic sub-units. Each sub-unit also follows the predicate-argument form.

For example, in Fig.~\ref{fig:casestudy}, we convert sentence $A$ into a hierarchical factorization tree $(a4)$ using a series of operations. The root node of the tree contains the semantic unit ``chase Tom Jerry little yard big'', which is the reordered representation of the original sentence ``The little Jerry is being chased by Tom in the big yard'' in a semantically normalized form. 
Moreover, the semantic unit at depth $0$ is factorized into four sub-units at depth $1$: ``chase'', ``Tom'', ``Jerry little'' and ``yard big'', each in the ``predicate-argument'' form. And at depth $2$, the semantic sub-unit ``Jerry little'' is further factorized into two sub-units ``Jerry'' and ``little''. Finally, a semantic unit that contains only one token (e.g., ``chase'' and ``Tom'' at depth $1$) can not be further decomposed. Therefore, it only has one child node at the next depth through self-duplication.

We can observe that each depth of the tree contains all the tokens (except meaningless ones) in the original sentence, but re-organizes these tokens into semantic units of different granularities. 


\subsection{Hierarchical Sentence Factorization}
\label{subsec:sf}

We now describe our detailed procedure to transform a natural language sentence to the desired factorization tree mentioned above.
Our Hierarchical Sentence Factorization algorithm mainly consists of five steps: 1) AMR parsing and alignment, 2) AMR purification, 3) index mapping, 4) node completion, and 5) node traversal. The latter four steps are illustrated in the example in Fig.~\ref{fig:casestudy} from left to right.

\begin{figure}[tb]
\centering
\includegraphics[width=2.7in]{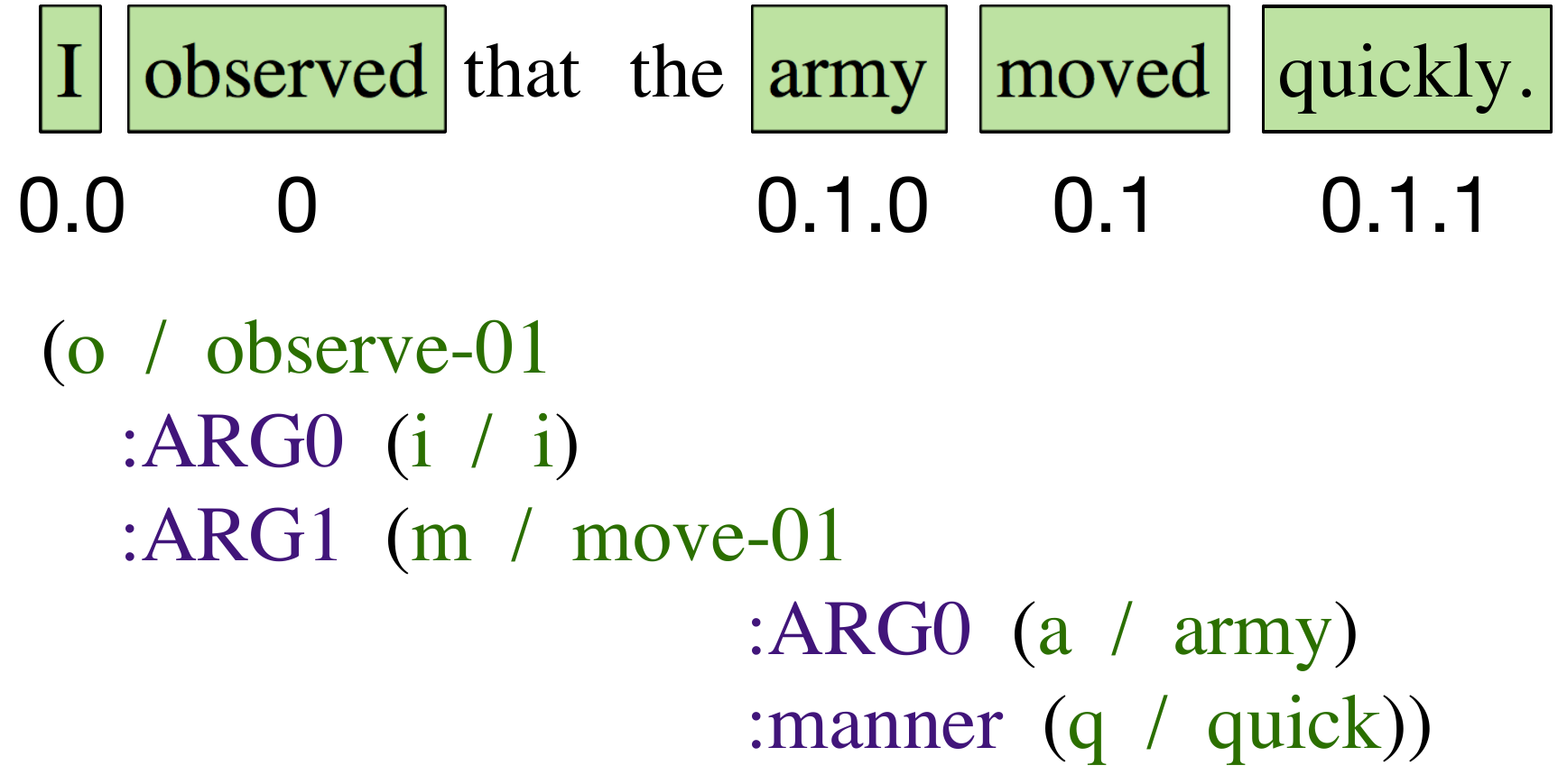}
\vspace{0mm}
\caption{An example of a sentence and its Abstract Meaning Representation (AMR), as well as the alignment between the words in the sentence and the nodes in AMR.}
\label{fig:amr}
\vspace{-3mm}
\end{figure}

\textbf{AMR parsing and alignment.}
Given an input sentence, the first step of our hierarchical sentence factorization algorithm is to acquire its Abstract Meaning Representation (AMR), as well as perform AMR-Sentence alignment to align the concepts in AMR with the tokens in the original sentence.
 
Semantic parsing \cite{baker1998berkeley,kingsbury2002treebank,berant2014semantic,banarescu2013abstract, damonte2016incremental} can be performed to generate the formal semantic representation of a sentence.
Abstract Meaning Representation (AMR) \cite{banarescu2013abstract} is a semantic parsing language that represents a sentence by a directed acyclic graph (DAG).
Each AMR graph can be converted into an AMR tree by duplicating the nodes that have more than one parent.

Fig.~\ref{fig:amr} shows the AMR of the sentence ``I observed that the army moved quickly.''
In an AMR graph, leaves are labeled with concepts, which represent either English words (e.g., ``army''), PropBank framesets (e.g., ``observe-01'') \cite{kingsbury2002treebank}, or special keywords (e.g., dates, quantities, world regions, etc.). For example, ``(a / army)'' refers to an instance of the concept army, where ``a'' is the variable name of army (each entity in AMR has a variable name).  ``ARG0'', ``ARG1'', ``:manner'' are different kinds of relations defined in AMR. Relations are used to link entities. For example, ``:manner'' links ``m / move-01'' and ``q / quick'', which means ``move in a quick manner''. Similarly, ``:ARG0'' links ``m / move-01'' and ``a / army'', which means that ``army'' is the first argument of ``move''.


Each leaf in AMR is a concept rather than the original token in a sentence. 
The alignment between a sentence and its AMR graph is not given in the AMR annotation. Therefore, AMR alignment \cite{pourdamghani2014aligning} needs to be performed to link the leaf nodes in the AMR to tokens in the original sentence.
Fig.~\ref{fig:amr} shows the alignment between sentence tokens and AMR concepts by the alignment indexes.
The alignment index $0$ is for the root node, 0.0 for the first child of the root node, 0.1 for the second child of the root node, and so forth. For example, in Fig.~\ref{fig:amr}, the word ``army'' in sentence is linked with index ``0.1.0'', which represents the concept node ``a / army'' in its AMR.
We refer interested readers to \cite{banarescu2013abstract,banarescu2012abstract} for more detailed description about AMR. 

Various parsers have been proposed for AMR parsing and alignment \cite{flanigan2014discriminative,wang2015boosting}. We choose the JAMR parser \cite{flanigan2014discriminative} in our algorithm implementation. 

\begin{figure}[tb]
\centering
\includegraphics[width=3.0in]{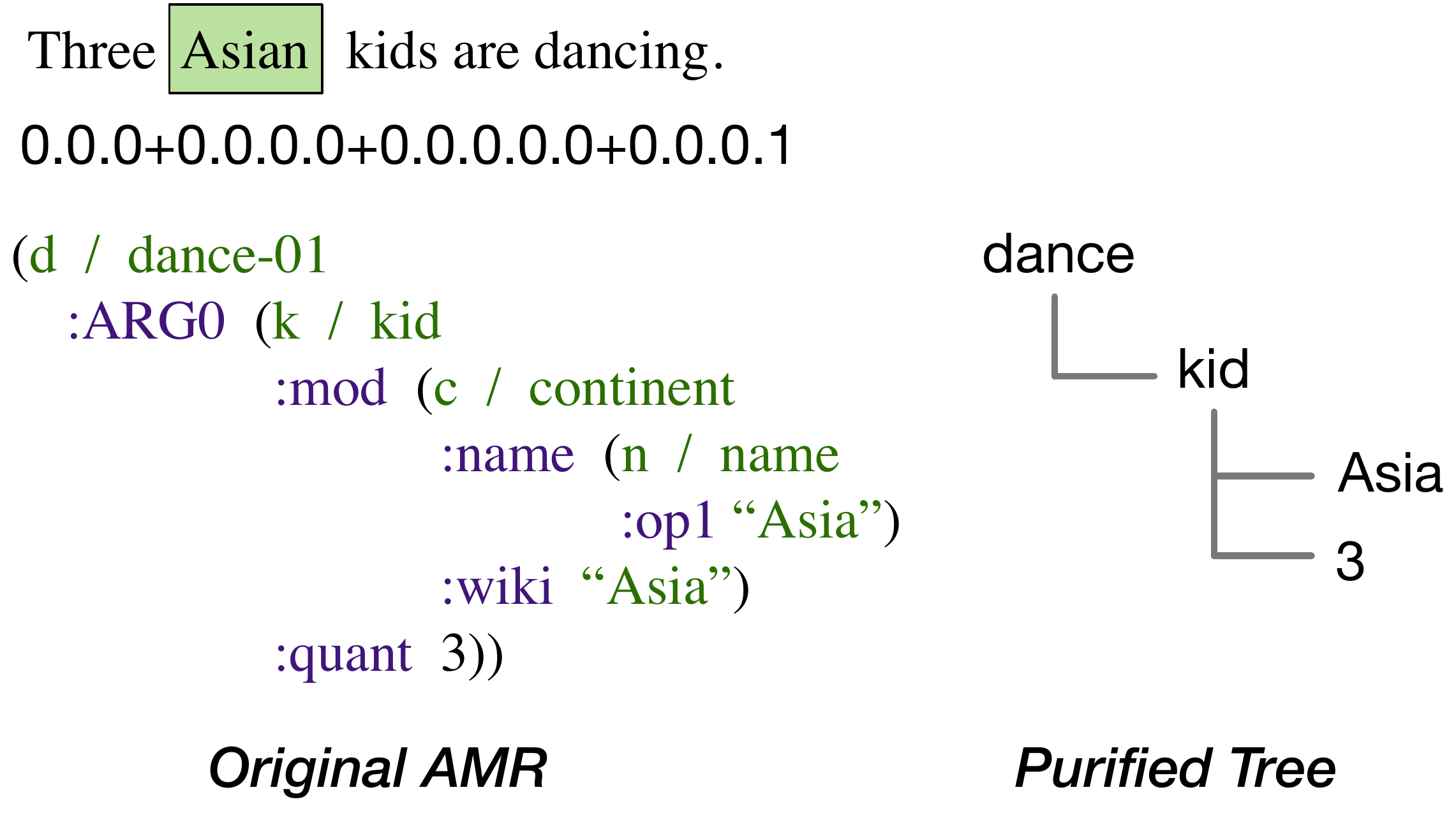}
\vspace{0mm}
\caption{An example to show the operation of AMR purification.}
\label{fig:amr2}
\vspace{-3mm}
\end{figure}

\textbf{AMR purification.} Unfortunately, AMR itself cannot be used to form the desired factorization tree. 
First, it is likely that multiple concepts in AMR may link to the same token in the sentence.
For example, Fig.~\ref{fig:amr2} shows AMR and its alignment for the sentence ``Three Asian kids are dancing.''.
The token ``Asian'' is linked to four concepts in the AMR graph: `` continent (0.0.0)'', ``name (0.0.0.0)'', ``Asia (0.0.0.0.0)'' and ``wiki Asia (0.0.0.1)''.
This is because AMR will match a named entity with predefined concepts which it belongs to, such as ``c / continent'' for ``Asia'', and form a compound representation of the entity. For example, in Fig.\ref{fig:amr2}, the token ``Asian'' is represented as a continent whose name is Asia, and its Wikipedia entity name is also Asia.

In this case, we select the link index with the smallest tree depth as the token's position in the tree.
Suppose $\mathcal{P}_w = \{p_1, p_2, \cdots, p_{|\mathcal{P}|}\}$ denotes the set of alignment indexes of token $w$.
We can get the desired alignment index of $w$ by calculating the longest common prefix of all the index strings in  $\mathcal{P}_w$.
After getting the alignment index for each token, we then replace the concepts in AMR with the tokens in sentence by the alignment indexes, and remove relation names (such as ``:ARG0'') in AMR, resulting into a compact tree representation of the original sentence, as shown in the right part of Fig.~\ref{fig:amr2}.

\textbf{Index mapping.}
A purified AMR tree for a sentence obtained in the previous step is still not in our desired form.
To transform it into a hierarchical sentence factorization tree, we perform index mapping and calculate a new position (or index) for each token in the desired factorization tree given its position (or index) in the purified AMR tree.
Fig.~\ref{fig:casestudy} illustrates the process of index mapping. After this step, for example, the purified AMR trees in Fig.~\ref{fig:casestudy} $(a1)$ and $(b1)$ will be transformed into $(a2)$ and $(b2)$.

Specifically, let $T^p_S$ denote a purified AMR tree of sentence $S$, and $T^f_S$ our desired sentence factorization tree of $S$.
Let $I^p_N = \overline{i_0.i_1.i_2.\cdots.i_d}$ denote the index of node $N$ in $T^p_S$, where $d$ is the depth of $N$ in $T^p_S$ (where depth 0 represents the root of a tree).
Then, the index $I^f_N$ of node $N$ in our desired factorization tree $T^f_S$ will be calculated as follows:
\begin{equation}
\label{eqn:transform-index}
I^f_N := \begin{cases}
 		     \overline{0.0} & \quad \text{if } d=0, \\ 
 			 \overline{i_0.(i_1+1).(i_2+1).\cdots.(i_d + 1)} & \quad \text{otherwise}. 
		 \end{cases}
\end{equation}
After index mapping, we add an empty root node with index $0$ in the new factorization tree, and link all nodes at depth $1$ to it as its child nodes. Note that the $i_0$ in every node index will always be 0.

\textbf{Node completion.}
We then perform node completion to make sure each branch of the factorization tree have the same maximum depth and to fill in the missing nodes caused by index mapping, illustrated by Fig.~\ref{fig:casestudy} $(a3)$ and $(b3)$.

First, given a pre-defined maximum depth $D$, for each leaf node $N^l$ with depth $d < D$ in the current $T^f_S$ after index mapping, we duplicate it for $D - d$ times and append all of them sequentially to $N^l$, as shown in Fig.~\ref{fig:casestudy} $(a3)$, $(b3)$, such that the depths of the ending nodes will always be $D$. For example, in Fig.~\ref{fig:casestudy} with $D = 2$, the node ``chase (0.0)'' and ``Tom (0.1)'' will be extended to reach depth $2$ via self-duplication.

Second, after index mapping, the children of all the non-leaf nodes, except the root node, will be indexed starting from 1 rather than 0. For example, in Fig.~\ref{fig:casestudy} $(a2)$, the first child node of ``Jerry (0.2)'' is ``little (0.2.1)''. 
In this case, we duplicate ``Jerry (0.2)'' itself to ``Jerry (0.2.0)'' to fill in the missing first child of ``Jerry (0.2)''. Similar filling operations are done for other non-leaf nodes after index mapping as well.

\textbf{Node traversal to complete semantic units}.
Finally, we complete each semantic unit in the formed factorization tree via node traversal, as shown in Fig.~\ref{fig:casestudy} $(a4)$, $(b4)$.
For each non-leaf node $N$, we traverse its sub-tree by Depth First Search (DFS). The original semantic unit in $N$ will then be replaced by the concatenation of the semantic units of all the nodes in the sub-tree rooted at $N$, following the order of traversal.

For example, for sentence $A$ in Fig.~\ref{fig:casestudy}, after node traversal, the root node of the factorization tree becomes ``chase Tom Jerry little yard big'' with index ``0''. We can see that the original sentence has been reordered into a predicate-argument structure. A similar structure is generated for the other nodes at different depths. 
Until now, each depth of the factorization tree $T^f_S$ can express the full sentence $S$  in terms of  semantic units at different granularity.

\section{Ordered Word Mover's Distance}
\label{sec:owmd}

\begin{figure}[tb]
\centering
\includegraphics[width=2.5in]{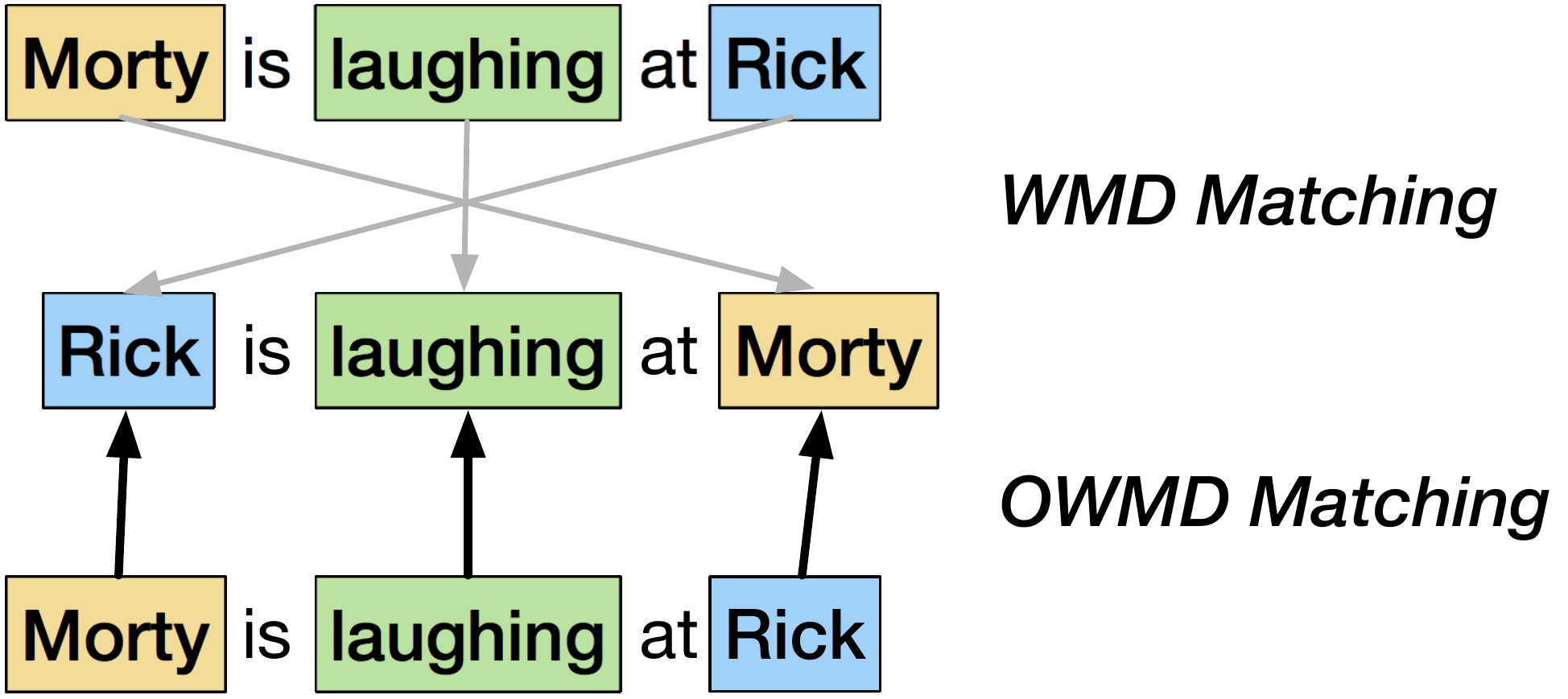}
\vspace{0mm}
\caption{Compare the sentence matching results given by Word Mover's Distance and Ordered Word Mover's Distance.}
\label{fig:match}
\vspace{-3mm}
\end{figure}

The proposed hierarchical sentence factorization technique naturally reorders an input sentence into a unified format at the root node. In this section, we introduce the \textit{Ordered Word Mover's Distance} metric which measures the semantic distance between two input sentences based on the unified representation of reordered sentences.

Assume $X\in \mathbb{R}^{d\times n}$ is a \textit{word2vec} embedding matrix for a vocabulary of $n$ words, and the $i$-th column $\mathbf{x}_i \in \mathbb{R}^d$ represents the $d$-dimensional embedding vector of $i$-th word in vocabulary.
Denote a sentence $S = \overline{a_1a_2\cdots a_K}$ where $a_i$ represents the $i$-th word (or the word embedding vector).
The Word Mover's Distance considers a sentence $S$ as its normalized bag-of-words (nBOW) vectors where the weights of the words in $S$ is $\mathbf \alpha = \{\alpha_1, \alpha_2, \cdots, \alpha_K\}$. Specifically, if word $a_i$ appears $c_i$ times in $S$, then $\alpha_i = \frac{c_i}{\sum_{j=1}^K c_j}$.

The Word Mover's Distance metric combines the normalized bag-of-words representation of sentences with Wasserstein distance (also known as Earth Mover's Distance \cite{rubner2000earth}) to measure the semantic distance between two sentences. 
Given a pair of sentences $S_1 = \overline{a_1a_2\cdots a_M}$ and $S_2 = \overline{b_1b_2\cdots b_N}$, where 
$b_j \in \mathbb{R}^d$ is the embedding vector of the $j$-th word in $S_2$. Let $\mathbf \alpha = \{\alpha_1, \cdots, \alpha_M\}$ and $\mathbf \beta = \{\beta_1, \cdots, \beta_N\}$ represents the normalized bag-of-words vectors of $S_1$ and $S_2$. We can calculate a distance matrix $D \in \mathbb{R}^{M\times N}$ where each element
$D_{ij} = \|a_i - b_j\|_2$ measures the distance between word $a_i$ and $b_j$ (we use the same notation to denote the word itself or its word vector representation).
Let $T \in \mathbb{R}^{M\times N}$ be a non-negative sparse transport matrix where $T_{ij}$ denotes the portion of word $a_i \in S_1$ that transports to word $b_j \in S_2$.
The Word Mover's Distance between sentences $S_1$ and $S_2$ is given by $\sum_{i,j} T_{ij} D_{ij}$. The transport matrix $T$ is computed solving the following constrained optimization problem:
\begin{equation}
\label{eq:wmd}
\begin{split}
	\underset{T \in \mathbb{R}_{+}^{M\times N}}{\mbox{minimize}}\quad 		& \sum_{i,j} T_{ij} D_{ij} \\
	\mbox{subject to}\quad & \sum\limits_{i = 1}^{M}  T_{ij} = \beta_j \quad 1\leq j \leq N,\\
			   & \sum\limits_{j = 1}^{N}  T_{ij} = \alpha_i \quad 1\leq i \leq M.
\end{split}
\end{equation}
Where the minimum ``word travel cost'' between two bags of words for a pair of sentences is calculated to measure the their semantic distance.

However, the Word Mover's Distance fails to consider a few aspects of natural language. First, it omits the sequential structure. For example, in Fig.~\ref{fig:match}, the pair of sentences ``Morty is laughing at Rick'' and ``Rick is laughing at Morty'' only differ in the order of words. The Word Mover's Distance metric will then find an exact match between the two sentences and estimate the semantic distance as zero, which is obviously false. Second, the normalized bag-of-words representation of a sentence can not distinguish duplicated words shown in multiple positions of a sentence.

To overcome the above challenges, we propose a new kind of semantic distance metric named Ordered Word Mover's Distance (OWMD). The Ordered Word Mover's Distance combines our sentence factorization technique with Order-preserving Wasserstein Distance proposed in \cite{su2017order}. It casts the calculation of semantic distance between texts as an optimal transport problem while preserving the sequential structure of words in sentences. The Ordered Word Mover's Distance differs from the Word Mover's Distance in multiple aspects.

First, rather than using normalized bag-of-words vector to represent a sentence, we decompose and re-organize a sentence using the sentence factorization algorithm described in Sec.~\ref{sec:sentence}. Given a sentence $S$, we represent it by the reordered word sequence $S'$ in the root node of its sentence factorization tree. Such representation normalizes a sentence into ``predicate-argument'' structure to better handle syntactic variations.
For example, after performing sentence factorization, sentences ``Tom is chasing Jerry'' and ``Jerry is being chased by Tom'' will both be normalized as ``chase Tom Jerry''.

Second, we calculate a new transport matrix $T$ by solving the following optimization problem
\begin{equation}
\label{eq:owmd}
\begin{split}
	\underset{T \in \mathbb{R}_{+}^{M\times N}}{\mbox{minimize}}\quad 		& \sum_{i,j} T_{ij} D_{ij} - \lambda_1 I(T) + \lambda_2 KL(T||P)\\
	\mbox{subject to}\quad & \sum\limits_{i = 1}^{M}  T_{ij} = \beta_j' \quad 1\leq j \leq N',\\
			   & \sum\limits_{j = 1}^{N}  T_{ij} = \alpha_i' \quad 1\leq i \leq M',
\end{split}
\end{equation}
where $\lambda_1 > 0$ and $\lambda_2 > 0$ are two hyper parameters.
$M'$ and $N'$ denotes the number of words in $S_1'$ and $S_2'$.
$\alpha_i'$ denotes the weight of the $i$-th word in normalized sentence $S_1'$ and $\beta_j'$ denotes the weight of the $j$-th word in normalized sentence $S_2'$. Usually we can set $\mathbf{\alpha'} = (\frac{1}{M'}, \cdots, \frac{1}{M'})$ and $\mathbf{\beta'} = (\frac{1}{N'}, \cdots, \frac{1}{N'})$ without any prior knowledge of word differences.

The first penalty term $I(T)$ is the inverse difference moment \cite{albregtsen2008statistical} of the transport matrix $T$ that measures local homogeneity of $T$. It is defined as:
\begin{equation}
\label{eq:IT}
\begin{split}
	I(T) = \sum\limits_{i=1}^{M'} \sum\limits_{j=1}^{N'} \frac{T_{ij}}{(\frac{i}{M'} - \frac{j}{N'})^2 + 1}.
\end{split}
\end{equation}
$I(T)$ will have a relatively large value if the large values of $T$ mainly appear near its diagonal.

Another penalty term $KL(T||P)$ denotes the KL-divergence between $T$ and $P$. 
$P$ is a two-dimensional distribution used as the prior distribution for values in $T$. It is defined as
\begin{equation}
\label{eq:P}
\begin{split}
	P_{ij} = \frac{1}{\sigma \sqrt{2\pi}}e^{- \frac{l^2(i,j)}{2\sigma^2}}
\end{split}
\end{equation}
where $l(i, j)$ is the distance from position $(i, j)$ to the diagonal line, which is calculated as
\begin{equation}
\label{eq:l}
\begin{split}
	l(i, j) = \frac{|i/M' - j/N'|}{\sqrt{1/M'^2 + 1/N'^2}}.
\end{split}
\end{equation}
As we can see, the farther a word in one sentence is from the other word in another sentence in terms of word orders, the less likely it will be transported to that word. Therefore, by introducing the two penalty terms $I(T)$ and $KL(T||P)$ into problem~(\ref{eq:owmd}), 
we encourage words at similar positions in two sentences to be matched.
Words at distant positions are less likely to be matched by $T$.

The problem (\ref{eq:owmd}) has a unique optimal solution $T^{\lambda_1, \lambda_2}$ since both the objective and the feasible set are convex. It has been proved in \cite{su2017order} that the optimal $T^{\lambda_1, \lambda_2}$ has the same form with $diag(\mathbf{k}_1) \cdot \mathbf{K} \cdot diag(\mathbf{k}_2)$, where $diag(\mathbf{k}_1) \in \mathbb{R}^{M'}$ and $diag(\mathbf{k}_2) \in \mathbb{R}^{N'}$ are two diagonal matrices with strictly positive diagonal elements. $\mathbf{K} \in \mathbb{R}^{M'\times N'}$ is a matrix defined as
\begin{equation}
\label{eq:K}
\begin{split}
	K_{ij} = P_{ij} e^{\frac{1}{\lambda_2}(S_{ij}^{\lambda_1} - D_{ij})},
\end{split}
\end{equation}
where
\begin{equation}
\label{eq:S}
\begin{split}
	S_{ij} = \frac{\lambda_1}{(\frac{i}{M'} - \frac{j}{N'})^2 + 1}.
\end{split}
\end{equation}
The two matrices $\mathbf{k}_1$ and $\mathbf{k}_2$ can be efficiently obtained by the Sinkhorn-Knopp iterative matrix scaling algorithm \cite{knight2008sinkhorn}:
\begin{equation}
\label{eq:k1}
\begin{split}
	\mathbf{k}_1 &\leftarrow \mathbf{\alpha'} ./ K \mathbf{k}_2, \\
	\mathbf{k}_2 &\leftarrow \mathbf{\beta'} ./ K^T \mathbf{k}_1.
\end{split}
\end{equation}
where $./$ is the element-wise division operation.
Compared with Word Mover's Distance, the Ordered Word Mover's Distance 
considers the positions of words in a sentence,
and is able to distinguish duplicated words at different locations. For example, in Fig.~\ref{fig:match}, while the WMD finds an exact match and get a semantic distance of zero for the sentence pair ``Morty is laughing at Rick'' and ``Rick is laughing at Morty'', the OWMD metric is able to find a better match relying on the penalty terms, and gives a semantic distance greater than zero.

The computational complexity of OWMD is also effectively reduced compared to WMD. With the additional constraints, the time complexity is $O(dM'N')$ where $d$ is the dimension of word vectors \cite{su2017order}, while it is $O(dp^3\log p)$ for WMD, where $p$ denotes the number of uniques words in sentences or documents \cite{kusner2015word}.

\section{Multi-scale Sentence Matching}
\label{sec:multi-layer}

\begin{figure*}[!htb]
\centering
\includegraphics[width=5.5in]{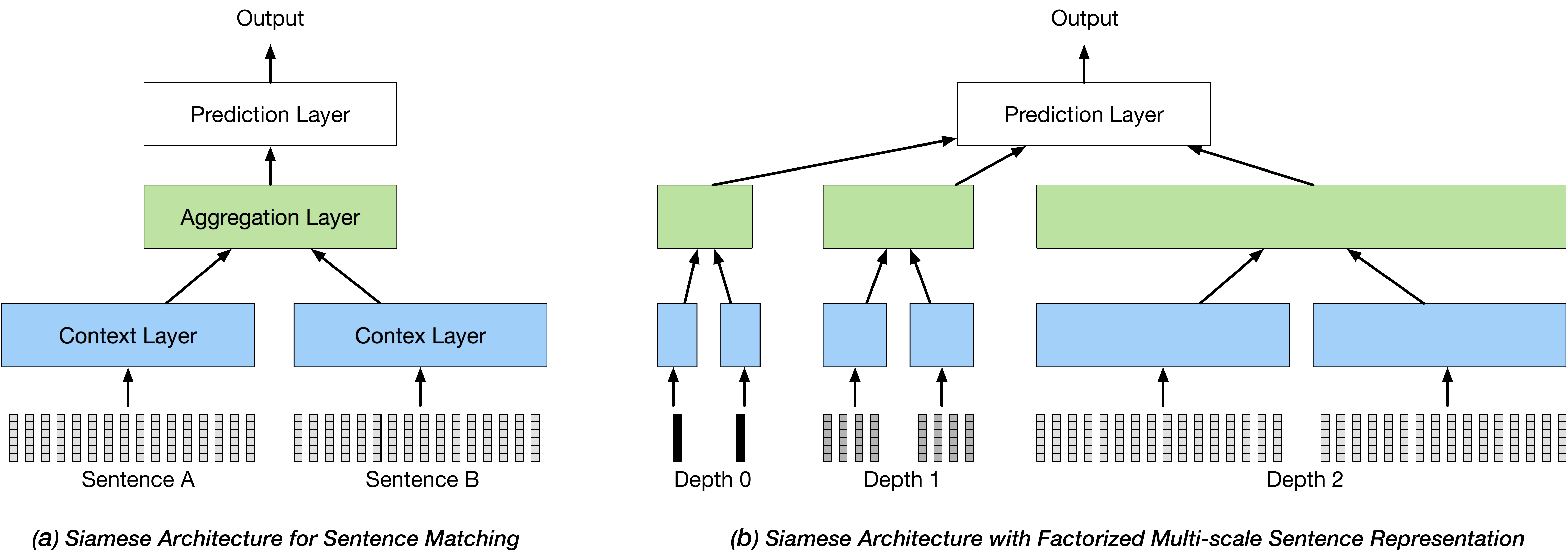}
\vspace{0mm}
\caption{Extend the Siamese network architecture for sentence matching by feeding into the multi-scale representations of sentence pairs.}
\label{fig:network}
\vspace{-2mm}
\end{figure*}

Our sentence factorization algorithm parses a sentence $S$ into a hierarchical factorization tree $T^f_S$, where each depth of $T^f_S$ contains the semantic units of the sentence at a different granularity.
In this section, we exploit this multi-scaled representation of $S$ present in $T^f_S$ to propose a multi-scaled Siamese network architecture that can extend any existing CNN or RNN-based Siamese architectures to leverage the hierarchical representation of sentence semantics.

Fig.~\ref{fig:network} (a) shows the network architecture of the popular Siamese ``matching-aggregation'' framework \cite{wang2016compare,mueller2016siamese,severyn2015learning,neculoiu2016learning,baudivs2016sentence} for sentence matching tasks. The matching process is usually performed as follows: First, the sequence of word embeddings in two sentences will be encoded by a context representation layer, which usually contains one or multiple layers of LSTM, bi-directional LSTM (BiLSTM), or CNN with max pooling layers.
The goal is to capture the contextual information of each sentence into a context vector. 
In a Siamese network, every sentence is encoded by the same context representation layer.
Second, the context vectors of two sentences will be concatenated in the aggregation layer. They may be further transformed by more layers of neural network
to get a fixed length matching vector.
Finally, a prediction layer will take in the matching vector and outputs a similarity score for the two sentences or the probability distribution over different sentence-pair relationships.

Compared with the typical Siamese network shown in Fig.~\ref{fig:network} (a), our proposed architecture shown in Fig.~\ref{fig:network} (b) differs in two aspects.
First, our network contains three Siamese sub-modules that are similar to (a). They correspond to the factorized representations from depth $0$ (the root layer) to depth $2$. We only select the semantic units from the top $3$ depths of the factorization tree as our input, because usually most semantic units at depth $2$ are already single words and can not be further factorized. Second, for each Siamese sub-module in our network architecture, the input is not the embedding vectors of words from the original sentences. Instead, we use semantic units at different depths of sentence factorization tree 
for matching.
We sum up the embedding vectors of the words contained in a semantic unit to represent that unit. Assuming each semantic unit at depth $d$ can be factorized into $k$ semantic sub-units at depth $d + 1$. If a semantic unit has less than $k$ sub-units, we add empty units as its child node to make each non-leaf node in a factorization tree has exactly $k$ child nodes. The empty units are embedded with a vector of zeros. After this procedure, the number of semantic units at depth $d$ of a sentence factorization tree is $k^d$.

Taking Fig.~\ref{fig:casestudy} as an example. We set $k = 4$ in Fig.~\ref{fig:casestudy}. For sentence A ``The little Jerry is being chased by Tom in the big yard'', the input at depth $0$ is the sum of word embedding $\{$chase, Tom, Jerry, little, yard, big$\}$. The input at depth $1$ are the embedding vectors of four semantic units: 
$\{$chase, Tome, Jerry little, yard big$\}$. Finally, at depth $2$, the semantic units are $\{$chase, -, -, -, Tom, -, -, -, Jerry, little, -, -, yard, big, -, -$\}$, where ``$-$'' denotes an empty unit.

As we can see, based on this factorized sentence representation, our network architecture explicitly matches a pair of sentences at several semantic granularities. 
In addition, we align the semantic units in two sentences by mapping their positions in the tree to the corresponding indices in the input layer of the neural network.
For example, as shown in Fig.~\ref{fig:casestudy}, the semantic units at depth $2$ are aligned according to their unit indices: ``chase'' matches with ``catch'', ``Tom'' matches with ``cat blue'', ``Jerry little'' matches with ``mouse brown'', and ``yard big'' matches with ``forecourt''.

\section{Evaluation}
\label{sec:eval}

In this section, we evaluate the performance of our unsupervised Ordered Word Mover's Distance metric and supervised Multi-scale Sentence Matching model with factorized sentences as input. We apply our algorithms to semantic textual similarity estimation tasks and sentence pair paraphrase identification tasks, based on four datasets: STSbenchmark, SICK, MSRP and MSRvid. 

\subsection{Experimental Setup}
\label{subsec:setup}

\begin{table}[tb]
  \caption{Description of evaluation datasets.}
  \label{tab:datasets}
  \begin{tabular}{lllll}
    \toprule
    Dataset & Task & Train & Dev & Test\\
    \midrule
    STSbenchmark & Similarity scoring & $5748$ & $1500$ & $1378$ \\
    SICK & Similarity scoring & $4500$ & $500$ & $4927$ \\
    MSRP & Paraphrase identification & $4076$ & - & $1725$ \\
    MSRvid & Similarity scoring & $750$ & - & $750$ \\
    \bottomrule
  \end{tabular}
  \vspace{-2mm}
\end{table}

We will start with a brief description for each dataset:
\begin{itemize}
\item \textbf{STSbenchmark}\cite{cer2017semeval}: it is a dataset for semantic textual similarity (STS) estimation. The task is to assign a similarity score to each sentence pair on a scale of 0.0 to 5.0, with 5.0 being the most similar.

\item \textbf{SICK}\cite{marelli2014sick}: it is another STS dataset from the SemEval 2014 task 1. It has the same scoring mechanism as STSbenchmark, where 0.0 represents the least amount of relatedness and 5.0 represents the most.

\item \textbf{MSRvid}: the Microsoft Research Video Description Corpus contains 1500 sentences that are concise summaries on the content of a short video. Each pair of sentences is also assigned a semantic similarity score between 0.0 and 5.0. 

\item \textbf{MSRP}\cite{quirk2004monolingual}: the Microsoft Research Paraphrase Corpus is a set of 5800 sentence pairs collected from news articles on the Internet. Each sentence pair is labeled 0 or 1, with 1 indicating that the two sentences are paraphrases of each other.
\end{itemize}

Table \ref{tab:datasets} shows a detailed breakdown of the datasets used in evaluation.
For STSbenchmark dataset we use the provided train/dev/test split.
The SICK dataset does not provide development set out of the box, so we extracted 500 instances from the training set as the development set.
For MSRP and MSRvid, since their sizes are relatively small to begin with, we did not create any development set for them.

One metric we used to evaluate the performance of our proposed models on the task of semantic textual similarity estimation is the Pearson Correlation coefficient, commonly denoted by $r$. Pearson Correlation is defined as:
\begin{equation}
\label{eq:pearson}
 r = cov(X,Y) /( \sigma_X \sigma_Y),
\end{equation}
where $cov(X,Y)$ is the co-variance between distributions X and Y, and $\sigma_X$, $\sigma_Y$ are the standard deviations of X and Y.
The Pearson Correlation coefficient can be thought as a measure of how well two distributions fit on a straight line. Its value has range [-1, 1], where a value of 1 indicates that data points from two distribution lie on the same line with a positive slope.

Another metric we utilized is the Spearman's Rank Correlation coefficient. Commonly denoted by $r_s$, the Spearman's Rank Correlation coefficient shares a similar mathematical expression with the Pearson Correlation coefficient, but it is applied to ranked variables.
Formally it is defined as \cite{wiki:spearman}:
\begin{equation}
\label{eq:spearman}
 \rho = cov(rg_X, rg_Y) / (\sigma_{rg_X} \sigma_{rg_Y}),
\end{equation}
where $rg_X$, $rg_Y$ denotes the ranked variables derived from $X$ and $Y$. $cov(rg_X,rg_Y)$, $\sigma_{rg_X}$, $\sigma_{rg_Y}$ corresponds to the co-variance and standard deviations of the rank variables. The term ranked simply means that each instance in X is ranked higher or lower against every other instances in X and the same for Y. We then compare the rank values of X and Y with \ref{eq:spearman}. Like the Pearson Correlation coefficient, the Spearman's Rank Correlation coefficient has an output range of [-1, 1], and it measures the monotonic relationship between X and Y. A Spearman's Rank Correlation value of 1 implies that as X increases, Y is guaranteed to increase as well.
The Spearman's Rank Correlation is also less sensitive to noise created by outliers compared to the Pearson Correlation.

For the task of paraphrase identification, the classification accuracy of label $1$ and the F1 score are used as metrics. 

In the supervised learning portion, we conduct the experiments on the aforementioned four datasets. We use training sets to train the models, development set to tune the hyper-parameters and each test set is only used once in the final evaluation. For datasets without any development set, we will use cross-validation in the training process to prevent overfitting, that is, use $10\%$ of the training data for validation and the rest is used in training. For each model, we carry out training for 10 epochs. We then choose the model with the best validation performance to be evaluated on the test set.

\subsection{Unsupervised Matching with OWMD}
\label{subsec:eval-owmd}

To evaluate the effectiveness of our Ordered Word Mover's Distance metric, we first take an unsupervised approach towards the similarity estimation task on the STSbenchmark, SICK and MSRvid datasets. Using the distance metrics listed in Table \ref{tab:compare-pearson} and \ref{tab:compare-spearman}, we first computed the distance between two sentences, then calculated the Pearson Correlation coefficients and the Spearman's Rank Correlation coefficients between all pair's distances and their labeled scores. We did not use the MSRP dataset since it is a binary classification problem.

In our proposed Ordered Word Mover's Distance metric, distance between two sentences is calculated using the order preserving Word Mover's Distance algorithm. For all three datasets, we performed hyper-parameter tuning using the training set and calculated the Pearson Correlation coefficients on the test and development set. We found that for the STSbenchmark dataset, setting $\lambda_1=10$, $\lambda_2=0.03$ produces the most optimal result. For the SICK dataset, a combination of $\lambda_1=3.5$, $\lambda_2=0.015$ works best. And for the MSRvid dataset, the highest Pearson Correlation is attained when $\lambda_1=0.01$, $\lambda_2=0.02$.
We maintain a max iteration of 20 since in our experiments we found that it is sufficient for the correlation result to converge.
During hyper-parameter tuning we discovered that using the Euclidean metric along with $\sigma=10$ produces better results, so all OWMD results summarized in Table \ref{tab:compare-pearson} and \ref{tab:compare-spearman} are acquired under these parameter settings. Finally, it is worth mentioning that our OWMD metric calculates the distances using factorized versions of sentences, while all other metrics use the original sentences. Sentence factorization is a necessary preprocessing step for the OWMD metric.

We compared the performance of Ordered Word Mover's Distance metric with the following methods:

\begin{itemize}
\item \textbf{Bag-of-Words (BoW)}: in the Bag-of-Words metric, distance between two sentences is computed as the cosine similarity between the word counts of the sentences.

\item \textbf{LexVec}~\cite{salle2016enhancing}: calculate the cosine similarity between the  averaged 300-dimensional LexVec word embedding of the two sentences. 

\item \textbf{GloVe}~\cite{pennington2014glove}: calculate the cosine similarity between the averaged 300-dimensional GloVe 6B word embedding of the two sentences. 

\item \textbf{Fastext}~\cite{joulin2016bag}: calculate the cosine similarity between the averaged 300-dimensional Fastext word embedding of the two sentences. 

\item \textbf{Word2vec}~\cite{mikolov2013efficient}: calculate the cosine similarity between the averaged 300-dimensional Word2vec word embedding of the two sentences.

\item \textbf{Word Mover's Distance (WMD)}~\cite{kusner2015word}: estimating the semantic distance between two sentences by WMD introduced in Sec.~\ref{sec:owmd}.
\end{itemize}

\begin{table}[tb]
  \caption{Pearson Correlation results on different distance metrics.}
  \label{tab:compare-pearson}
  \begin{tabular}{c|cc|cc|c}
    \toprule
    \multirow{2}{*}{Algorithm} & \multicolumn{2}{c}{STSbenchmark} & \multicolumn{2}{c}{SICK} & MSRvid\\ 
     & Test & Dev & Test & Dev & Test\\
    \midrule
    BoW & $0.5705$ & $0.6561$ & $0.6114$ & $0.6087$ & $0.5044$ \\
    LexVec & $0.5759$ & $0.6852$ & $0.6948$ & $\mathbf{0.6811}$ & $0.7318$\\
    GloVe & $0.4064$ & $0.5207$ & $0.6297$ & $0.5892$  & $0.5481$ \\
    Fastext & $0.5079$ & $0.6247$ & $0.6517$ & $0.6421$  & $0.5517$  \\
    Word2vec & $0.5550$ & $0.6911$ & $\mathbf{0.7021}$ & $0.6730$  & $0.7209$  \\
    WMD & $0.4241$ & $0.5679$ & $0.5962$ & $0.5953$  & $0.3430$  \\
    OWMD & $\mathbf{0.6144}$ & $\mathbf{0.7240}$ & $0.6797$ & $0.6772$  & $\mathbf{0.7519}$  \\
    \bottomrule
  \end{tabular}
  \vspace{-4mm}
\end{table}

\begin{table}[tb]
  \caption{Spearman's Rank Correlation results on different distance metrics.}
  \label{tab:compare-spearman}
  \begin{tabular}{c|cc|cc|c}
    \toprule
    \multirow{2}{*}{Algorithm} & \multicolumn{2}{c}{STSbenchmark} & \multicolumn{2}{c}{SICK} & MSRvid\\ 
     & Test & Dev & Test & Dev & Test\\
    \midrule
    BoW & $0.5592$ & $0.6572$ & $0.5727$ & $0.5894$ & $0.5233$ \\
    LexVec & $0.5472$ & $0.7032$ & $0.5872$ & $0.5879$ & $0.7311$\\
    GloVe & $0.4268$ & $0.5862$ & $0.5505$ & $0.5490$  & $0.5828$ \\
    Fastext & $0.4874$ & $0.6424$ & $0.5739$ & $0.5941$  & $0.5634$  \\
    Word2vec & $0.5184$ & $0.7021$ & $0.6082$ & $0.6056$  & $0.7175$  \\
    WMD & $0.4270$ & $0.5781$ & $0.5488$ & $0.5612$  & $0.3699$  \\
    OWMD & $\mathbf{0.5855}$ & $\mathbf{0.7253}$ & $\mathbf{0.6133}$ & $\mathbf{0.6188}$  & $\mathbf{0.7543}$  \\
    \bottomrule
  \end{tabular}
  \vspace{-2mm}
\end{table}

Table \ref{tab:compare-pearson} and Table \ref{tab:compare-spearman} compare the performance of different metrics in terms of the Pearson Correlation coefficients and the Spearman's Rank Correlation coefficients.
We can see that the result of our OWMD metric achieves the best performance on all the datasets in terms of the Spearman's Rank Correlation coefficients.
It also produced the best Pearson Correlation results on the STSbenchmark and the MSRvid dataset, while the performance on SICK datasets are close to the best.
This can be attributed to the two characteristics of OWMD. First, the input sentence is re-organized into a predicate-argument structure using the sentence factorization tree. Therefore, corresponding semantic units in the two sentences will be aligned roughly in order. Second, our OWMD metric takes word positions into consideration and penalizes disordered matches. Therefore, it will produce less mismatches compared with the WMD metric.



\subsection{Supervised Multi-scale Semantic Matching}
\label{subsec:eval-multilayer}

\begin{table*}[tb]
  \caption{A comparison among different supervised learning models in terms of accuracy, F1 score, Pearson's $r$ and Spearman's $\rho$ on various test sets.}
  \label{tab:sts}
  \begin{tabular}{c|cc|cc|cc|cc}
    \toprule
    \multirow{2}{*}{Model} & \multicolumn{2}{c}{MSRP} & \multicolumn{2}{c}{SICK} & \multicolumn{2}{c}{MSRvid} & \multicolumn{2}{c}{STSbenchmark}\\ 
     & Acc.(\%) & F1(\%) & $r$ & $\rho$ & $r$ & $\rho$ & $r$ & $\rho$ \\
    \midrule
    MaLSTM & $66.95$ & $73.95$ & $0.7824$ & $0.71843$ & $0.7325$ & $0.7193$ & $0.5739$ & $0.5558$\\
    Multi-scale MaLSTM & $\mathbf{74.09}$ & $\mathbf{82.18}$ & $\mathbf{0.8168}$ & $\mathbf{0.74226}$ & $\mathbf{0.8236}$ & $\mathbf{0.8188}$ & $\mathbf{0.6839}$ & $\mathbf{0.6575}$\\
    \midrule
    HCTI & $73.80$ & $80.85$ & $0.8408$ & $0.7698$ & $\mathbf{0.8848}$ & $\mathbf{0.8763}$  & $\mathbf{0.7697}$ & $\mathbf{0.7549}$ \\
    Multi-scale HCTI & $\mathbf{74.03}$ & $\mathbf{81.76}$ & $\mathbf{0.8437}$ & $\mathbf{0.7729}$ & $0.8763$ & $0.8686$  & $0.7269$ & $0.7033$  \\
    \bottomrule
  \end{tabular}
  \vspace{-2mm}
\end{table*}

The use of sentence factorization can improve both existing unsupervised metrics and existing supervised models. 
To evaluate how the performance of existing Siamese neural networks can be improved by our sentence factorization technique and the multi-scale Siamese architecture, we implemented two types of Siamese sentence matching models, HCTI \cite{mueller2016siamese} and MaLSTM \cite{shao2017hcti}. HCTI is a Convolutional Neural Network (CNN) based Siamese model, which achieves the best Pearson Correlation coefficient on STSbenchmark dataset in SemEval2017 competition (compared with all the other neural network models). MaLSTM is a Siamese adaptation of the Long Short-Term Memory (LSTM) network for learning sentence similarity. As the source code of HCTI is not released in public, we implemented it according to \cite{shao2017hcti} by Keras \cite{chollet2015keras}. With the same parameter settings listed in paper \cite{shao2017hcti} and tried our best to optimize the model, we got a Pearson correlation of 0.7697 (0.7833 in paper \cite{shao2017hcti}) in STSbencmark test dataset.

We extended HCTI and MaLSTM to our proposed Siamese architecture in Fig. \ref{fig:network}, namely the Multi-scale MaLSTM and the Multi-scale HCTI. To evaluate the performance of our models, the experiment is conducted on two tasks: 1) semantic textual similarity estimation based on the STSbenchmark, MSRvid, and SICK2014 datasets; 2) paraphrase identification based on the MSRP dataset.

Table \ref{tab:sts} shows the results of HCTI, MaLSTM and our multi-scale models on different datasets. Compared with the original models, our models with multi-scale semantic units of the input sentences as network inputs significantly improved the performance on most datasets. 
Furthermore, the improvements on different tasks and datasets also proved the general applicability of our proposed architecture.

Compared with MaLSTM, our multi-scaled Siamese models with factorized sentences as input perform much better on each dataset. For MSRvid and STSbenmark dataset, both Pearson's $r$ and Spearman's $\rho$ increase about $10\%$ with Multi-scale MaLSTM. Moreover, the Multi-scale MaLSTM achieves the highest accuracy and F1 score on the MSRP dataset compared with other models listed in Table \ref{tab:sts}.

There are two reasons why our Multi-scale MaLSTM significantly outperforms MaLSTM model. First, for an input sentence pair, 
we explicitly model their semantic units with the factorization algorithm.
Second, our multi-scaled network architecture is 
specifically designed
for multi-scaled sentences representations. Therefore, it is able to explicitly match a pair of sentences at different granularities.

We also report the results of HCTI and Multi-scale HCTI in Table \ref{tab:sts}. For the paraphrase identification task, our model shows better accuracy and F1 score on MSRP dataset. For the semantic textual similarity estimation task, the performance varies across datasets. On the SICK dataset, the performance of Multi-scale HCTI is close to HCTI with slightly better Pearson' $r$ and Spearman's $\rho$. However, the Multi-scale HCTI is not able to outperform HCTI on MSRvid and STSbenchmark. HCTI is still the best neural network model on the STSbenchmark dataset, and the MSRvid dataset is a subset of STSbenchmark.
Although HCTI has strong performance on these two datasets, it performs worse than our model on other datasets.
Overall, the experimental results demonstrated the general applicability of our proposed model architecture, which performs well on various semantic matching tasks.

\section{Related Work}
\label{sec:related}

The task of natural language sentence matching has been extensively studied for a long time. Here we review related unsupervised and supervised models for sentence matching.

Traditional unsupervised metrics for document representation, including bag of words (BOW), term frequency inverse document frequency (TF-IDF) \cite{wu2008interpreting}, Okapi BM25 score \cite{robertson1994some}. However, these representations can not capture the semantic distance between individual words.
Topic modeling approaches such as Latent Semantic Indexing (LSI) \cite{deerwester1990indexing} and Latent Dirichlet Allocation (LDA) \cite{blei2003latent} attempt to circumvent the problem through learning a latent representation of documents.
But when applied to semantic-distance based tasks such as text-pair semantic similarity estimation, these algorithms usually cannot achieve good performance.

Learning distributional representation for words, sentences or documents based on deep learning models have been popular recently. \textit{word2vec} \cite{mikolov2013efficient} and \textit{Glove} \cite{pennington2014glove} are two high quality word embeddings that have been extensively used in many NLP tasks. Based on word vector representation, the Word Mover's Distance (WMD) \cite{kusner2015word} algorithm measures the dissimilarity between two sentences (or documents)
as the minimum distance that
the embedded words of one sentence need to
``travel'' to reach the embedded words of another
sentence. However, when applying these approaches to sentence pair matching tasks, the interactions between sentence pairs are omitted, also the ordered and hierarchical structure of natural languages is not considered.

Different neural network architectures have been proposed for sentence pair matching tasks. Models based on Siamese architectures \cite{mueller2016siamese,severyn2015learning,neculoiu2016learning,baudivs2016sentence} usually transform the word embedding sequences of text pairs into context representation vectors through a multi-layer Long Short-Term Memory (LSTM) \cite{sundermeyer2012lstm} network or Convolutional Neural Networks (CNN) \cite{krizhevsky2012imagenet}, followed by a fully connected network or score function which gives the similarity score or classification label based on the context representation vectors. However, Siamese models defer the interaction between two sentences until the hidden representation layer, therefore may lose details of sentence pairs for matching tasks \cite{hu2014convolutional}.

Aside from Siamese architectures, \cite{wang2017bilateral} introduced a matching layer into Siamese network to compare the contextual embedding of one sentence with another. \cite{hu2014convolutional,pang2016text} proposed convolutional matching models that consider all pair-wise interactions between words in sentence pairs.  \cite{he2016pairwise} propose to explicitly model pairwise word interactions with a pairwise word interaction similarity cube and a similarity focus layer to identify important word interactions.

\section{Conclusion}
\label{sec:conclude}

In this paper, we propose a technique named \textit{Hierarchical Sentence Factorization} that is able to transform a sentence into a hierarchical factorization tree. Each node in the tree is a semantic unit consists of one or several words in the sentence and reorganized into the form of ``predicate-argument'' structure. Each depth in the tree factorizes the sentence into semantic units of different scales.
Based on the hierarchical tree-structured representation of sentences, we propose both an unsupervised metric and two supervised deep models for sentence matching tasks. On one hand, we design a new unsupervised distance metric, named \textit{Ordered Word Mover's Distance} (OWMD), to measure the semantic difference between a pair of text snippets.
OWMD takes the sequential structure of sentences into account, and is able to handle the flexible syntactical structure of natural language sentences.
On the other hand, we propose the multi-scale Siamese neural network architecture which takes the multi-scale representation of a pair of sentences as network input and matches the two sentences at different granularities.

We apply our techniques to the task of text-pair similarity estimation and the task of text-pair paraphrase identification, based on multiple datasets. Our extensive experiments show that both the unsupervised distance metric and the supervised multi-scale Siamese network architecture can achieve significant improvement on multiple datasets using the technique of sentence factorization.

\bibliographystyle{ACM-Reference-Format}
\balance
\bibliography{main} 

\end{document}